\documentclass[conference]{IEEEtran}

\IEEEoverridecommandlockouts

\usepackage{cite}
\usepackage{amsmath,amssymb,amsfonts}
\usepackage{algorithmic}
\usepackage{graphicx}
\usepackage{textcomp}
\usepackage{xcolor}
\usepackage{multirow}
\usepackage[normalem]{ulem}

\def\BibTeX{{\rm B\kern-.05em{\sc i\kern-.025em b}\kern-.08em
    T\kern-.1667em\lower.7ex\hbox{E}\kern-.125emX}}
\begin{document}

\title{RMES: Real-Time Micro-Expression Spotting Using Phase From Riesz Pyramid

\thanks{This work was sponsored in part by the Hong Kong Innovation and Technology Fund (grant ITS/210/20), Bright Nation Limited and the Hong Kong Research Grants Council (grant 16214821).}
}

\author{
    \IEEEauthorblockN{\textbf{Yini Fang}\IEEEauthorrefmark{1}\IEEEauthorrefmark{2}, \textbf{Didan Deng}\IEEEauthorrefmark{1}, \textbf{Liang Wu}\IEEEauthorrefmark{1}, \textbf{Frederic Jumelle}\IEEEauthorrefmark{4}\IEEEauthorrefmark{2}, and 
    \textbf{Bertram Shi}\IEEEauthorrefmark{1}
    }
    \IEEEauthorblockA{\IEEEauthorrefmark{1}Hong Kong University of Science and Technology}
    \IEEEauthorblockA{\IEEEauthorrefmark{2}Ydentity Organization}
    \IEEEauthorblockA{\IEEEauthorrefmark{4}Bright Nation Limited}

    \IEEEauthorblockA{\textit {\{yfangba, ddeng, liang.wu\}@connect.ust.hk, f.jumelle@brightnationlimited.com, eebert@ust.hk} }}

\maketitle

\begin{abstract}
Micro-expressions (MEs) are involuntary and subtle facial expressions that are thought to reveal feelings people are trying to hide. ME spotting detects the temporal intervals containing MEs in videos. Detecting such quick and subtle motions from long videos is difficult. Recent works leverage detailed facial motion representations, such as the optical flow, and deep learning models, leading to high computational complexity. To reduce computational complexity and achieve real-time operation,  we propose RMES, a real-time ME spotting framework. We represent motion using phase computed by Riesz Pyramid, and feed this motion representation into a three-stream shallow CNN, which predicts the likelihood of each frame belonging to an ME. In comparison to optical flow, phase provides more localized motion estimates, which are essential for ME spotting, resulting in higher performance. Using phase also reduces the required computation of the ME spotting pipeline by 77.8\%. Despite its relative simplicity and low computational complexity, our framework achieves state-of-the-art performance on two public datasets: CAS(ME)$^2$ and SAMM Long Videos. 
\end{abstract}


\section{Introduction}
Macro-expressions are the commonly encountered facial expressions like happiness, sadness, surprise or anger. They typically last from 1/2 to 4 seconds and are often generated consciously or voluntarily. Micro-expressions (ME) are much shorter in time (typically 1/25 to 1/2 second) and smaller in magnitude \cite{yan2013fast}. They are thought to be largely involuntary, making them difficult to hide, even when people seek to suppress them. They are essential non-verbal communication clues, providing insight into the genuine emotional state of individuals. They have been widely to various fields such as public safety, recruitment, medicine, and neuropsychology \cite{li2017towards}.

ME spotting is an important first step in automated facial expression recognition (AFER) \cite{duque2018micro}. 
An ME has three distinct stages: onset, apex, and offset. The onset occurs when the facial muscles start to contract. The apex is the facial action's point of peak intensity. The offset marks the moment when muscles return to their neutral state. ME spotting aims to detect onset and offset of all MEs in a video sequence. 

Because MEs are composed of quick and subtle motions with low spatial amplitude, the best-performing ME spotting algorithms usually extract image motion as optical flow at the front end, followed by deep neural network models performing detection at the back end. However, the high computational complexity of optical flow makes them difficult to run in real time, especially in resource-constrained environments, such as remote monitoring for public safety. Thus, researchers seek to reduce computational complexity while maintaining high accuracy. Liong et al. \cite{liong2021shallow} proposed to reduce back end complexity with a three-stream shallow Convolutional neural network (CNN), while using optical flow as the front end. Other work has focused on the front end, using phase as an alternative to the optical flow. Phase is often used as a pre-processing step for more complex image motion algorithms, such as motion magnification and optical flow estimation. It provides local motion estimates, which more complex optical flow algorithms refine by integrating information across space, e.g. using smoothness constraints, to address issues like the aperture problem.

In this paper, we propose the RMES (\textbf{R}eal-time \textbf{M}icro-\textbf{E}xpression \textbf{S}potting) framework, which reduces computational complexity at both ends, combining phase features extracted by a Riesz Pyramid in the front end with a lightweight three-stream shallow CNN in the back end.  Our approach uses a shallow CNN structure similar to that of Liong et al. \cite{liong2021shallow}, but with phase instead of optical flow to represent motion information. 

To the best of our knowledge, this paper describes the first comparison of phase versus optical flow for a ME-related task.  With the changes to the network mentioned above and detailed below, using phase instead of optical flow improves the F1 score by 26.94\%, as we demonstrate in Sec. \ref{abl}. Although optical flow reduces noise and solves the aperture problem, it also removes the small variations in local motion that are critical to detecting MEs. However, to use phase effectively, an additional facial alignment step is critical. It is also important to choose the scale from the Riesz pyramid correctly, as there is a trade-off between motion sensitivity and motion range, as we describe in Sec. \ref{rmesfr}.

As an added benefit, phase reduces inference time by 77.8\% compared to optical flow. Despite its simplicity and low computational complexity, our method achieves the state-of-the-art F1 scores on both the CAS(ME)$^2$ and SAMM Long Video datasets.

\section{Related Work}

Image motion information is critical for ME spotting, requiring integration of information over both time and space. Some approaches extract spatial features first, then integrate them over time. For example, 
Yang et al. \cite{yang2021facial}  present a deep learning framework based on facial action units (AU) detection in each frame, followed by the combination of AU information across time. Other approaches integrate across both dimensions simultaneously. For example, Yap et al. \cite{yap20223d} propose an end-to-end 3D-CNN framework that learns features directly from appearance.

Most recent works in ME spotting use explicit representations of image motion, such as optical flow. This is then fed into a classifier, such as a deep CNN model \cite{wang2021mesnet, yu2021lssnet, liong2022mtsn, liong2023spot}. This combination achieves the state-of-the-art results in expression spotting. However, it is time-consuming, as optical flow extraction and deep CNNs are both computationally complex. Liong et al. \cite{liong2021shallow} addressed this partially with a shallow three-stream CNN model to predict the likelihood score for both macro- and micro-expressions from optical flow. This reduced the time complexity of the neural network, but not the time complexity of optical flow. 

Optical flow is the most common representation for motion information. A notable exception is Duque et al.'s work \cite{duque2018micro}\cite{duque2020mean}, which we discuss in more detail below. Although optical flow is intuitively appealing and widely used, we argue that it is not the best motion representation for ME spotting. Optical flow algorithms usually integrate information over large spatial regions, under the assumption that the flow is smooth, i.e., values at neighbouring pixels are close to each other. This is a good assumption for rigid objects, and helps to address the aperture problem, which arises because local measurements can only estimate displacements orthogonal to the dominant texture orientation. 
However, it is a poor assumption for the face, which is non-rigid, and for MEs, which are spatially localized. 

Given the small image displacements in MEs, some work has used motion magnification as a preprocessing step to amplify them. For example, Kumar et al. \cite{kumar2022three} first amplified motion using  Eulerian Motion Magnification (EMM), then computed optical flow and input it to a graph attention network. However, this approach is wasteful, as motion information is extracted twice: first by the EMM algorithm to magnify the motion and second by the optical flow algorithm to measure it. 

We argue that it is more efficient to use the motion information extracted during motion magnification directly. The most effective algorithms for motion magnification represent motion information using phase, which we describe in more detail in the next section. For example, Wadhwa et al.'s real-time phase-based motion magnification algorithm computes phase using the Riesz Pyramid \cite{wadhwa2014riesz}. Our proposed RMES algorithm uses the same representation. While phase is noisier than optical flow due to limited spatial integration, it provides more localized information, which can capture the non-rigid motion of the face during MEs.

Our approach differs from previous work using phase for facial emotion analysis. Deng et al.'s MiMaMo net for macro-expression emotion recognition used phase computed by a Complex Steerable Pyramid in the frequency domain \cite{deng2020mimamo}. We use the Riesz Pyramid, which is computed entirely in the spatial domain, making it easier to avoid artifacts due to phase wrap-around. It is also approximately four times faster. Duque et al. used the phase variance, a simple hand-crafted feature computed from Riesz Pyramid phase for ME spotting \cite{duque2018micro} and classification \cite{duque2020mean}. They spatially averaged the phase over three pre-defined facial regions (the two eyes and the mouth) to estimate the net motion in each region, then looked for peaks in the squared difference between the instantaneous phase and the mean phase over the entire video. This approach ignores spatially localized motion, e.g. in different parts of the eye or mouth, that is critical to ME spotting. It also ignores temporally localized motion, looking only at absolute changes in position versus the mean. In contrast, our approach preserves spatially and temporally localized motion information about motion. Rather than averaging, we maintain phase estimates at each pixel. Rather than measuring instantaneous versus global mean phase, we look at inter-frame phase differences.

\section{Phase Differences from the Riesz Pyramid}
\label{riez}

\begin{figure}[t]
\includegraphics[width=8.8cm]{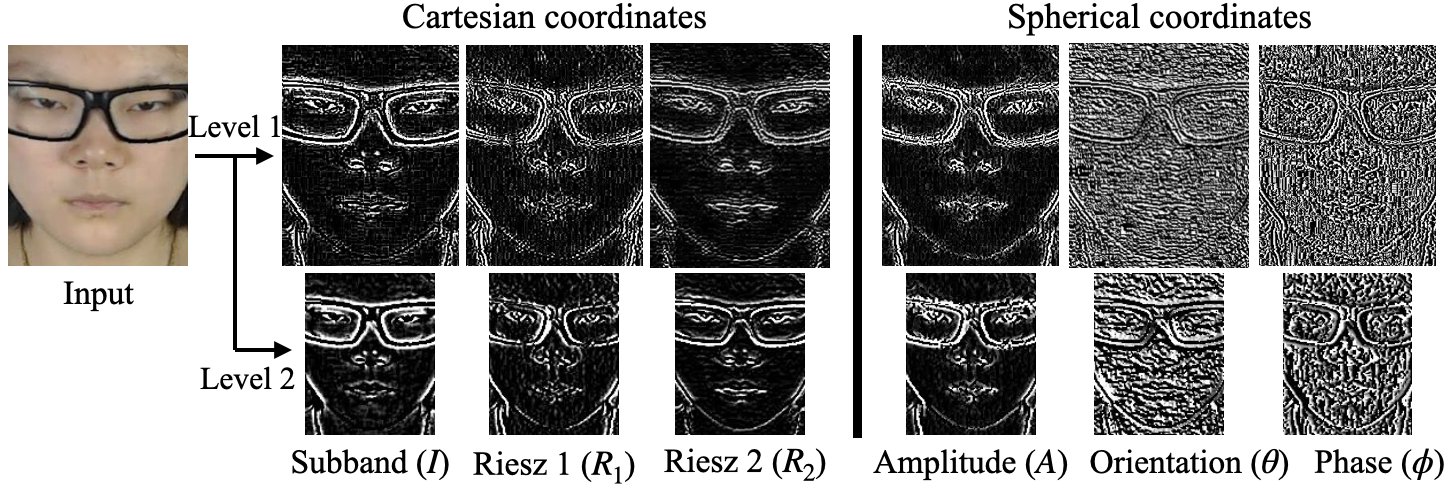}
\centering
\caption{Two representations of a 2-level Riesz Pyramid: Cartesian coordinates (i.e., monogenic signal) and spherical coordinates.
}
\label{pyramidExample}
\end{figure} 

Fig. \ref{pyramidExample} illustrates a two-level Riesz Pyramid constructed from an input face image. 

We first build a Laplacian pyramid for each image in the input video. Images at coarser levels are obtained from previous levels by Gaussian blurring and downsampling. We then calculate differences between the images at subsequent levels (subbands).  

Next, we take the approximate Riesz transform of each subband in the Laplacian pyramid. 
The Riesz transform is a steerable Hilbert transformer. We compute a quadrature pair of filters that are 90 degrees out of phase with respect to each other along the dominant orientation at every pixel \cite{wadhwa2014riesz}.
The filters have transfer functions:
$-i \frac{\omega_{x}}{\|\vec{\omega}\|} \text{ and } -i \frac{\omega_{y}}{\|\vec{\omega}\|}$.

Applying this filter pair to the image at frame $m$, $I_m$, we obtain a pair of filter responses, ($R_{1m}, R_{2m}$).  The input $I$ and the filter responses form a triple called the \textbf{monogenic signal} ($I_m, R_{1m}, R_{2m}$), which we combine into a quaternion $\mathbf{r}_m$, which can also be expressed in terms of the local amplitude $A_m$, local orientation $\theta_m$ and local phase $\phi_m$: 
\begin{align}
\mathbf{r}_m & = I_m + i R_{1m} + j R_{2m} \\
& = A_m \cos \phi_m + i  A_m \sin \phi_m \cos \theta_m \nonumber \\ 
& \ \ \ +  j A_m \sin \phi_m \sin \theta_m
\label{2}
\end{align}
The solution of Eq. \ref{2} is not unique, since both ($A_m, \phi_m, \theta_m$) and ($A_m, -\phi_m, \theta_m+\pi$) map to the same monogenic signal. Therefore, we use the \textbf{quaternionic phase} ($\phi_m \cos\theta_m, \phi_m \sin\theta_m$), which is invariant to the ambiguity between ($\phi_m, \theta_m$) and ($-\phi_m, \theta_m+\pi$). 

Following \cite{lee2002general} and \cite{wadhwa2014quaternionic}, 
we compute a sequence of quaternionic phase differences between adjacent frames from the sequence of unit quaternions 
$\mathbf{\hat{r}}_m = \mathbf{r}_m / \| \mathbf{r}_m \|$ according to
\begin{equation}
\label{eq:phasediff}
    \log (\mathbf{\hat{r}}_m \mathbf{\hat{r}}^{-1}_{m-1} )
    \approx i \Delta \phi_m \cos\theta_m
    + j \Delta \phi_m \sin\theta_m
\end{equation}
where the phase difference $\Delta \phi_m = \phi_m - \phi_{m-1}$ is a measure of the inter-frame motion in the direction of the local orientation $\theta_m$, which is measured with respect to the horizontal image axis. The approximation holds if the local orientation is roughly constant in time.

\section{RMES Framework}
\label{rmesfr}

\begin{figure}[t]
\includegraphics[width=8.5cm]{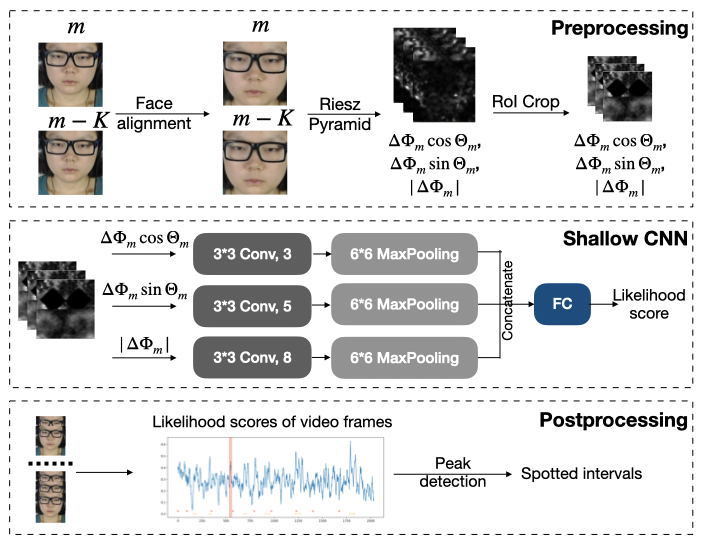}
\centering
\caption{The Real-time Micro-Expression Spotting (RMES) framework. }
\label{structure}
\end{figure} 

Fig. \ref{structure} shows the structure of our proposed RMES framework. Inspired by \cite{liong2021shallow}, the structure consists of three stages: preprocessing, shallow CNN, and postprocessing. 

In the preprocessing step, we apply face alignment and the Riesz Pyramid to obtain a sequence of quaternionic phase differences as described in the previous section. 

We then accumulate these phase differences over $K$ frames to measure the motion between images spaced $K$ frames apart. We choose $K$ to be half the average length of the MEs in the dataset. If the apex typically occurs about halfway between the onset and offset, then $K$ is the average number of frames from onset to apex and from apex to offset. Intuitively, the magnitude of facial movement should be largest for these intervals. 

In the shallow CNN stage, we crop signals from Regions of Interest (RoI) relevant to MEs and input them into the shallow CNN, which outputs a score representing the likelihood the frame belongs to an ME. This results in a sequence of $(T-K)$ scores, where $T$ is the video length. 

In postprocessing, we smooth the $(T-K)$ output scores and detect peaks above the predefined threshold as apex frames, $\{a_n\}_{n=1}^N$, where $N$ is the number of detected peaks. The final spotted intervals are $\{[a_n-K, a_n+K]\}_{n=1}^N$. 

We describe these steps in more detail below.

\textbf{Face alignment}. 
Face alignment is an essential preprocessing step to compensate for head motion. Prior approaches accounted for head motion by subtracting the average motion across the image or a region, such as the noise. While this can compensate for translation, it cannot compensate for other effects, such as tilt and rotation. Given the sensitivity of phase, our experimental results suggest that accurate face alignment is critical to ensuring that the Riesz Pyramid extracts motion relevant to ME. We do face alignment using the OpenFace Toolkit \cite{baltrusaitis2018openface}, which detects 68 facial landmarks and uses them to align the face by linear warping followed by cropping of the face region. We resize cropped and aligned images to $224\times224$ pixel resolution.

\textbf{Riesz Pyramid}.
\label{pyramidSec} 
We temporally filter the sequence of quaternionic phase differences in Eq. \ref{eq:phasediff} to isolate motions of interest and remove noise. Since we seek to detect subtle motions, we use a non-causal finite impulse response (FIR) filter with no group delay. Unlike EMM, which focuses on amplifying periodic motions, ME motions are non-periodic. Rather than using the temporal bandpass filter in EMM, which can result in large oscillations near the onset and offset of the ME due to the Gibbs phenomenon, we use a lowpass filter, which filters out high frequency noise.

We accumulate the filtered phase difference sequence to measure the motion between frames spaced $K$ frames apart. If, with a slight abuse of notation, we use $i \Delta \phi_m \cos \theta_m + j \Delta \phi_m \sin \theta_m$ to denote the filtered sequence, the accumulated sequence is given by $
i \Delta \Phi_m \cos \Theta_m + j \Delta \Phi_m \sin \Theta_m $ where
\begin{align}
    \Delta \Phi_m \cos \Theta_m & = \sum_{k=0}^{K-1} \Delta \phi_{m-k} \cos \theta_{m-k}  \label{eq:acc1} \\
    \Delta \Phi_m \sin \Theta_m & = \sum_{k=0}^{K-1} \Delta \phi_{m-k} \sin \theta_{m-k}  \label{eq:acc2}
\end{align}
We also compute the phase difference magnitude
\begin{equation}
|\Delta\Phi_m| = 
\sqrt{
\left(\Delta\Phi_m \sin \Theta_m  \right)^{2}
+
\left( \Delta \Phi_m \cos \Theta_m \right)^{2}
}.
\end{equation}

The accumulation of the quaternionic phase differences in Eqs. \ref{eq:acc1} and \ref{eq:acc2} avoids phase overlap that might be encountered due to the relatively large displacements over $K$ frames, but still assumes that the phase difference between successive frames is always in the interval $(-\pi, \pi]$ \cite{wadhwa2014quaternionic}. This assumption is often valid for ME spotting, since small motions result in small phase differences. However, the size of the phase shift also depends upon the dominant spatial frequency, which varies with the level of the Laplacian pyramid. For the same motion, larger spatial frequencies (finer scales) result in the larger phase shifts. This increases sensitivity, but also increases the chance of phase wrap-around. Thus, the choice of pyramid level is a trade-off between sensitivity and motion range.

\textbf{RoI cropping}. 
We further crop the images based on the Regions of Interest (RoIs) where MEs might occur according to Facial Action Coding System (FACS), i.e., the eyebrows and the mouth as illustrated in Fig. \ref{example} \cite{zhang2020spatio}. We then resample each region to $15 \times 30$ pixels and stack them, resulting in an input feature map with $30 \times 30$ pixels and three channels
($\Delta \Phi_m \cos \Theta_m$,
$\Delta \Phi_m \sin \Theta_m$ and
$|\Delta\Phi_m|$). 
We normalize the feature maps using the Z score. 

\textbf{Shallow CNN}. 
The shallow CNN has three streams: one for each channel. Each stream consists of a CNN layer (filter size $3\times3$  and stride size $1\times1$), followed by max pooling layer (kernel size $6\times6$  and stride size $6\times6$).
Following \cite{liong2021shallow}, the number of filters varies according to stream depending upon the importance of the features (3, 5, and 8 filters, respectively). For example, $\Delta \Phi_m \cos\Theta_m$ and $\Delta \Phi_m \sin\Theta_m$ measure horizontal and vertical displacements, respectively. Since MEs typically involve more vertical than horizontal movements, we allocate more filters to the $\Delta \Phi \sin \Theta_m$ stream. We concatenate and flatten the feature maps, resulting in a $5\times5\times16 = 400$ dimensional feature vector. This is fed into two fully connected layers, the first one (400, 400) to summarize the features, and the second one (400, 1) to output the final score $s$, which indicates the likelihood that the frame is part of an ME interval. 

We assign binary ground truth scores $S_i$ to each frame $i$ according to the Intersection Over Union (IoU) between the interval between two frames used to compute the phase difference at frame $i$ and the nearest ME interval in the ground truth, i.e., $S_i = 1 \text{ if } \text{IoU} \geq 0.5, \text{else } S_i = 0$.

We train the model with a Mean Square Error (MSE) loss function: $
L = \frac{1}{N} \sum_{i=1}^{N}\left(s_{i}-S_{i}\right)^{2},
$ where $N$ is the total number of image pairs in the dataset.

\textbf{Postprocessing}. 
For each video, we smooth the scores by averaging across the $2K+1$ frames: 
\begin{equation}
\hat{s}_{i}=\frac{1}{2 K+1} \sum_{j=i-K}^{i+K} s_{j}, \text { for } i=K+1, \ldots, T-K,
\end{equation}
Then, we perform peak detection to find local maxima exceeding a predefined threshold $H$ and such that the horizontal distance between adjacent peaks is no less than $k$. We set 
\begin{equation}
H=\hat{s}_{\text {mean }}+h \times\left(\hat{s}_{\text {max }}-\hat{s}_{\text {mean }}\right),
\end{equation}
where $\hat{s}_{\text {mean }}$ and $\hat{s}_{\text {max }}$ are the mean and maximum scores over the entire video. $h \in [0, 1]$ is a hyperparameter controlling the threshold. For each detected peak, we obtain a spotted interval $[P-K, P+K]$, where $P$ is the peak location.

\section{Methodology}

\textbf{Datasets}. 
We evaluated our model on two public datasets: CAS(ME)$^2$ \cite{qu2016cas} and SAMM Long Videos \cite{davison2016samm, davison2018objective, yap2020samm}. Table \ref{data} summarizes the properties of these datasets.

\textbf{Metrics}. 
We compare model performance using the F1 score proposed in \cite{li2019spotting}. Let $X_i$ be the ground truth number of micro-expressions in video $i$, $Y_i$ be the number of spotted intervals, and $a_i$ be the number of true positives. A spotted interval is considered to be a true positive if its IoU with a ground truth interval exceeds 0.5. The F1 score is defined as 
\begin{equation*}
    \text{F1}=\frac{2 P R}{P+R}
\end{equation*}
where the precision $P$ and recall $R$ are given by 
\begin{equation*}
    R  =\frac{\sum_{i=1}^{V} a_{i}}{\sum_{i=1}^{V} X_{i}} \text{ and }
    P  = \frac{\sum_{i=1}^{V} a_{i}}{\sum_{i=1}^{V} Y_{i}} 
\end{equation*}

\textbf{Implementation} We implemented our framework using PyTorch 1.11.0, and trained on an NVIDIA A40-16Q GPU. We apply Leave-one-subject-out (LOSO) cross-validation in order to ensure all samples are evaluated. We used the third level of the Riesz Pyramid. The cutoff frequency of the lowpass filter was 10Hz, corresponding to a time constant of 100ms. The value of $K$ was 6 and 47 for CAS(ME)$^2$ and SAMM Long Videos, respectively, half the average length of MEs in the two datasets (200ms and 235ms). The value of $h$ used for peak detection was 0.7. For our experiments with optical flow, we used the Python library OpenCV 4.5.2 running on the same device as phase calculation.

\begin{table}
\small
\centering
\caption{Summary of Dataset Properties}
\label{data}
\begin{tabular}{|c|c|c|c|c|c|}
\hline
         & \#Vid. & \#Sub. & \#Samp. & FPS & Resolution \\ \hline
CAS(ME)$^2$ & 98     & 22       & 57      & 30  & $640\times480$    \\ \hline
SAMMLV   & 147    & 32       & 159     & 200 & $2040\times1088$  \\ \hline
\end{tabular}
\end{table}

\section{Experimental Results}

Table \ref{res} compares the F1 scores of our RMES framework with other SOTA systems. Duque et al. \cite{duque2018micro} perform peak detection on phase variance computed from the Riesz Pyramid. Yang et al. \cite{yang2021facial} and Yap et al. \cite{yap20223d} use end-to-end deep learning frameworks with appearance-based representation. The others [4][7]-[10] cascade an optical flow front-end with a CNN-based back-end. 

Our RMES framework outperforms all other SOTA systems listed. To see the effect of phase versus optical flow, we compare our results with those of Liong et al. \cite{liong2021shallow}, who use an optical flow front end followed by a shallow CNN similar to that used in our model. Our model with phase improves the F1 score by 26.94\% and 9.67\% for CAS(ME)$^2$ and SAMM Long Videos. This supports our claim that phase features are better representations of facial motions for ME spotting than optical flow. The F1 scores listed do not tend to increase over time because most systems, except \cite{liong2023spot}, focused on optimizing an overall F1 score evaluating both macro- and micro-expression spotting, rather than micro-expression spotting alone.

Table \ref{time} compares the time complexity of Liong et al. \cite{liong2021shallow} and RMES. The inference time of these two methods only takes 1.6\% and 6.7\% of the overall time. Thus, focusing on improving the preprocessing step is vital to improving speed, as it takes most of the time. Even including the extra face alignment step, which accounts for the bulk (0.018 seconds or 68\%) of the preprocessing time, the preprocessing time of RMES is only 22\% of \cite{liong2021shallow}, corresponding to more than four times speedup.

Table \ref{flop} lists the number of parameters and FLOPs for all models where that information was provided. Our model requires the fewest FLOPs. Compared to Liong et al. \cite{liong2021shallow}, our model also reduces the computational complexity at the CNN back end, because we use smaller receptive fields ($3 \times 3$ vs. $5 \times 5$) and a smaller input dimension ($30 \times 30$ vs. $42 \times 42$).

\begin{figure}[t]
\includegraphics[width=7.8cm]{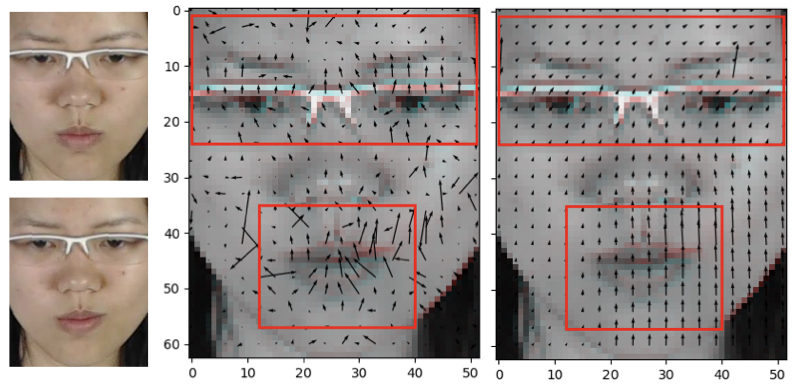}
\centering
\caption{Comparison of phase differences and the optical flow. Red rectangles show the RoI areas. (Left) An aligned image sequence pair. (Middle) Quiver plot of phase differences overlaid upon a red-green anaglyph of the image pair. (Right) Similar plot for optical flow.}
\label{example}
\end{figure}

Fig. \ref{example} compares the phase differences and optical flow for an aligned image sequence pair where the subject raises the right corner of her mouth. Due to the spatial integration in optical flow, its vector field is smoother and cleaner, but ignores the local details that phase preserves. For example, the non-rigid deformation of the mouth is reflected in the varying directions of the phase difference vectors, but the optical flow vectors all point in the same direction (upwards). These subtle changes in the face are vital for ME spotting and should not be smoothed out. This example supports our intuition that phase provides more robust and richer feature representation for micro-expression.

\begin{table}
\small
\centering
\caption{F1 scores of benchmarks and our model}
\label{res}
\begin{tabular}{|c|c|c|}
\hline
Methods      & CAS(ME)$^2$ & SAMM LV \\ \hline
Duque et al., 2018  \cite{duque2018micro} & 0.0806   &     0.0711      \\ \hline
Liong et al., 2021 \cite{liong2021shallow} & 0.1173   & 0.1520           \\ \hline
Wang et al., 2021 \cite{wang2021mesnet} & 0.0360 & 0.0880 \\ \hline
Yu et al., 2021 \cite{yu2021lssnet} & 0.0420 & 0.1310 \\ \hline
Yang et al., 2021 \cite{yang2021facial} & 0.0153 & 0.1155 \\ \hline
Liong et al., 2022 \cite{liong2022mtsn} & 0.0808 & 0.0878 \\ \hline
Yap et al., 2022 \cite{yap20223d} & 0.0714 & 0.0466 \\ \hline
Liong et al., 2023 \cite{liong2023spot} & 0.1214 & 0.0949 \\ \hline
Ours         & \textbf{0.1489}   & \textbf{0.1667} \\ \hline
\end{tabular}
\end{table}

\begin{table}
\small
\centering
\caption{Time complexity comparison (unit: second)}
\label{time}
\begin{tabular}{|c|c|c|c|}
\hline
Methods      & Preprocessing                            & Inference  & Overall  \\ \hline
Liong et al., 2021 \cite{liong2021shallow} & 0.12                                     & 0.0020 & 0.122     \\ \hline
Ours        & 0.0266 & 0.0019 & 0.0285    \\ \hline
\end{tabular}
\end{table}

\begin{table}[t]
\small
\centering
\caption{Model size and latency comparison}
\label{flop}
\begin{tabular}{|c|c|c|}
\hline
Methods      & \# Param                            & \# FLOPs   \\ \hline
Liong et al., 2021\cite{liong2021shallow} & 315k,                                     & 2.1M      \\ \hline
Yu et al., 2021\cite{yu2021lssnet}        & 23M & 810M     \\ \hline
Liong et al., 2022\cite{liong2022mtsn}        & 67k & 1.4M    \\ \hline
Liong et al., 2023\cite{liong2023spot}        & 161k & 2.7M    \\ \hline
Ours        & 161k & 0.6M     \\ \hline
\end{tabular}
 \vspace{-0.1in}
\end{table}

\section{Ablation Studies}
\label{abl}

We conducted ablation studies to show the effect of using face alignment, taking input from different levels of the Reisz pyramid, the choice of temporal filter in the pyramid, and cropping RoIs.

\textbf{Face alignment}. Face alignment (FA) reduces the effect of global translations, rotations and scaling, which are less relevant in ME spotting. Table \ref{fa} compares the effect of using/not using FA for both datasets. Using FA significantly reduces false positive detections, improving precision 
and therefore the F1 score by 21.7\% and 42.7\% on CAS(ME)$^2$ and SAMMLV, respectively. Without face alignment, global motion can introduce large phase differences, which the model classifies mistakenly as a ME.

\begin{table}[b]
\small
\centering
\caption{The effect of Face Alignment (FA). \\(TP = True Positive, FP = False Positive, FN = False Negative)}
\label{fa}
\begin{tabular}{|c|cccccc|}
\hline
\multirow{2}{*}{FA} & \multicolumn{6}{c|}{CAS(ME)$^2$}                                                                                                                         \\ \cline{2-7} 
                         & \multicolumn{1}{c|}{TP} & \multicolumn{1}{c|}{FP}  & \multicolumn{1}{c|}{FN}  & \multicolumn{1}{c|}{Recall} & \multicolumn{1}{c|}{Precision} & F1     \\ \hline
True               & \multicolumn{1}{c|}{14} & \multicolumn{1}{c|}{117} & \multicolumn{1}{c|}{43}  & \multicolumn{1}{c|}{0.2456} & \multicolumn{1}{c|}{0.1069}    & 0.1489 \\ \hline
False              & \multicolumn{1}{c|}{14} & \multicolumn{1}{c|}{158} & \multicolumn{1}{c|}{43}  & \multicolumn{1}{c|}{0.2456} & \multicolumn{1}{c|}{0.0814}    & 0.1223 \\ \hline
\multirow{2}{*}{FA} & \multicolumn{6}{c|}{SAMM Long Videos}                                                                                                                 \\ \cline{2-7} 
                         & \multicolumn{1}{c|}{TP} & \multicolumn{1}{c|}{FP}  & \multicolumn{1}{c|}{FN}  & \multicolumn{1}{c|}{Recall} & \multicolumn{1}{c|}{Precision} & F1     \\ \hline
True              & \multicolumn{1}{c|}{30} & \multicolumn{1}{c|}{171} & \multicolumn{1}{c|}{129} & \multicolumn{1}{c|}{0.1887} & \multicolumn{1}{c|}{0.1493}    & 0.1667 \\ \hline
False             & \multicolumn{1}{c|}{31} & \multicolumn{1}{c|}{341} & \multicolumn{1}{c|}{128} & \multicolumn{1}{c|}{0.195}  & \multicolumn{1}{c|}{0.0833}    & 0.1168 \\ \hline
\end{tabular}
\end{table}

\begin{figure}[t]
\includegraphics[width=8.8cm]{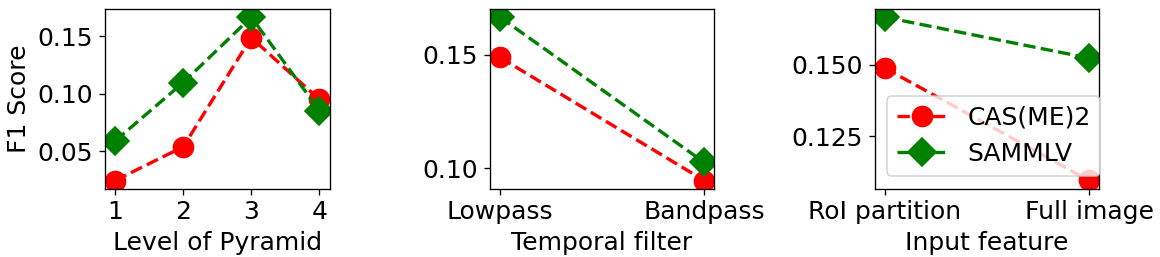}
\centering
\caption{Ablation studies on the effect of different pyramid levels, temporal filters, and use of RoIs.}
\label{ablation}
\end{figure}

\textbf{Riesz Pyramid Level}. Different levels in the Riesz Pyramid correspond to different frequency ranges or scales. Earlier levels correspond to finer scales and higher resolution. They contain more detailed information, but cannot represent large displacements, tend to be noisier and are more prone to phase wrap-around. The left plot in Fig. \ref{ablation} shows the F1 score when using phase from  different levels as input. The F1 score initially increases as we move up the pyramid, as displacement range increases and the noise decreases, but degrades after the third level due to the low resolution ($13\times13$) at level 4.

\textbf{Temporal Filter}. The middle plot in Fig. \ref{ablation} compares the effect of filtering the unwrapped phase differences with a lowpass filter with 10Hz cutoff frequency and a band-pass filter between 2Hz and 10Hz. Eliminating lower frequencies degrades performance.

\textbf{Region of Interest}. We compare the F1 score when using input only from eye and mouth RoIs and when using input from the full image. The right plot in Fig. \ref{ablation} shows that restricting input to RoIs results in better F1 scores. We designed the RoI based on Facial Action Coding System, which is used to describe MEs. This focuses the network on information relevant to MEs and avoids irrelevant information, such as from the jawline and areas below it.

\section{Conclusion}
In this paper, we addressed the high time complexity of ME spotting algorithms by proposing the Real-time Micro-Expression Spotting (RMES) framework, which uses phase extracted from Riesz Pyramid followed by a three-stream shallow CNN. In our evaluations on CAS(ME)$^2$ and SAMM Long Videos, RMES achieves the state-of-the-art performance with  significantly lower computational complexity. Our comparison of phase differences and optical flow suggests that phase differences are better suited for ME spotting, as it localized motion details. Moving forward, possible extensions of this work include replacing the peak detection postprocessing step with frame by frame labels, which will enable the framework to detect ME intervals with varying length. We also plan to explore approaches other than Gaussian smoothing for denoising, such as an attention mechanism that could dynamically assign small weights to noisy regions.

\bibliographystyle{IEEEbib}
\bibliography{reference}

\begin{thebibliography}{10}

\bibitem{yan2013fast}
Wen-Jing Yan, Qi~Wu, Jing Liang, Yu-Hsin Chen, and Xiaolan Fu,
\newblock ``How fast are the leaked facial expressions: {T}he duration of
  micro-expressions,''
\newblock {\em Journal of Nonverbal Behavior}, vol. 37, no. 4, pp. 217--230,
  2013.

\bibitem{li2017towards}
Xiaobai Li, Xiaopeng Hong, Antti Moilanen, Xiaohua Huang, Tomas Pfister,
  Guoying Zhao, and Matti Pietik{\"a}inen,
\newblock ``Towards reading hidden emotions: {A} comparative study of
  spontaneous micro-expression spotting and recognition methods,''
\newblock {\em IEEE Transactions on Affective Computing}, vol. 9, no. 4, pp.
  563--577, 2017.

\bibitem{duque2018micro}
Carlos~Arango Duque, Olivier Alata, Remi Emonet, Anne-Claire Legrand, and
  Hubert Konik,
\newblock ``Micro-expression spotting using the {R}iesz pyramid,''
\newblock in {\em 2018 IEEE Winter Conference on Applications of Computer
  Vision (WACV)}. IEEE, 2018, pp. 66--74.

\bibitem{liong2021shallow}
Gen-Bing Liong, John See, and Lai-Kuan Wong,
\newblock ``Shallow optical flow three-stream {CNN} for macro-and
  micro-expression spotting from long videos,''
\newblock in {\em 2021 IEEE International Conference on Image Processing
  (ICIP)}. IEEE, 2021, pp. 2643--2647.

\bibitem{yang2021facial}
Bo~Yang, Jianming Wu, Zhiguang Zhou, Megumi Komiya, Koki Kishimoto, Jianfeng
  Xu, Keisuke Nonaka, Toshiharu Horiuchi, Satoshi Komorita, Gen Hattori,
  et~al.,
\newblock ``Facial action unit-based deep learning framework for spotting
  macro-and micro-expressions in long video sequences,''
\newblock in {\em Proceedings of the 29th ACM International Conference on
  Multimedia}, 2021, pp. 4794--4798.

\bibitem{yap20223d}
Chuin~Hong Yap, Moi~Hoon Yap, Adrian Davison, Connah Kendrick, Jingting Li,
  Su-Jing Wang, and Ryan Cunningham,
\newblock ``3{D}-{CNN} for facial micro-and macro-expression spotting on long
  video sequences using temporal oriented reference frame,''
\newblock in {\em Proceedings of the 30th ACM International Conference on
  Multimedia}, 2022, pp. 7016--7020.

\bibitem{wang2021mesnet}
Su-Jing Wang, Ying He, Jingting Li, and Xiaolan Fu,
\newblock ``{MESN}et: A convolutional neural network for spotting multi-scale
  micro-expression intervals in long videos,''
\newblock {\em IEEE Transactions on Image Processing}, vol. 30, pp. 3956--3969,
  2021.

\bibitem{yu2021lssnet}
Wang-Wang Yu, Jingwen Jiang, and Yong-Jie Li,
\newblock ``{LSSN}et: A two-stream convolutional neural network for spotting
  macro-and micro-expression in long videos,''
\newblock in {\em Proceedings of the 29th ACM International Conference on
  Multimedia}, 2021, pp. 4745--4749.

\bibitem{liong2022mtsn}
Gen~Bing Liong, Sze-Teng Liong, John See, and Chee-Seng Chan,
\newblock ``{MTSN}: A multi-temporal stream network for spotting facial
  macro-and micro-expression with hard and soft pseudo-labels,''
\newblock in {\em Proceedings of the 2nd Workshop on Facial Micro-Expression:
  Advanced Techniques for Multi-Modal Facial Expression Analysis}, 2022, pp.
  3--10.

\bibitem{liong2023spot}
Gen-Bing Liong, John See, and Chee-Seng Chan,
\newblock ``Spot-then-{R}ecognize: A micro-expression analysis network for
  seamless evaluation of long videos,''
\newblock {\em Signal Processing: Image Communication}, vol. 110, pp. 116875,
  2023.

\bibitem{duque2020mean}
Carlos~Arango Duque, Olivier Alata, R{\'e}mi Emonet, Hubert Konik, and
  Anne-Claire Legrand,
\newblock ``Mean oriented {R}iesz features for micro expression
  classification,''
\newblock {\em Pattern Recognition Letters}, vol. 135, pp. 382--389, 2020.

\bibitem{kumar2022three}
Ankith Jain~Rakesh Kumar and Bir Bhanu,
\newblock ``Three stream graph attention network using dynamic patch selection
  for the classification of micro-expressions,''
\newblock in {\em Proceedings of the IEEE/CVF Conference on Computer Vision and
  Pattern Recognition}, 2022, pp. 2476--2485.

\bibitem{wadhwa2014riesz}
Neal Wadhwa, Michael Rubinstein, Fr{\'e}do Durand, and William~T Freeman,
\newblock ``Riesz pyramids for fast phase-based video magnification,''
\newblock in {\em 2014 IEEE International Conference on Computational
  Photography (ICCP)}. IEEE, 2014, pp. 1--10.

\bibitem{deng2020mimamo}
Didan Deng, Zhaokang Chen, Yuqian Zhou, and Bertram Shi,
\newblock ``{MIMAMO} {N}et: {I}ntegrating micro-and macro-motion for video
  emotion recognition,''
\newblock in {\em Proceedings of the AAAI Conference on Artificial
  Intelligence}, 2020, vol.~34, pp. 2621--2628.

\bibitem{lee2002general}
Jehee Lee and Sung~Yong Shin,
\newblock ``General construction of time-domain filters for orientation data,''
\newblock {\em IEEE Transactions on Visualization and Computer Graphics}, vol.
  8, no. 2, pp. 119--128, 2002.

\bibitem{wadhwa2014quaternionic}
Neal Wadhwa, Michael Rubinstein, Fr{\'e}do Durand, and William~T. Freeman,
\newblock ``Quaternionic representation of the {R}iesz pyramid for video
  magnification,''
\newblock Tech. {R}ep. 2014-009, MIT CSAIL, 2014.

\bibitem{baltrusaitis2018openface}
Tadas Baltrusaitis, Amir Zadeh, Yao~Chong Lim, and Louis-Philippe Morency,
\newblock ``Openface 2.0: {F}acial behavior analysis toolkit,''
\newblock in {\em 2018 13th IEEE International Conference on Automatic Face \&
  Gesture Recognition (FG 2018)}. IEEE, 2018, pp. 59--66.

\bibitem{zhang2020spatio}
Li-Wei Zhang, Jingting Li, Su-Jing Wang, Xian-Hua Duan, Wen-Jing Yan, Hai-Yong
  Xie, and Shu-Cheng Huang,
\newblock ``Spatio-temporal fusion for macro-and micro-expression spotting in
  long video sequences,''
\newblock in {\em 2020 15th IEEE International Conference on Automatic Face and
  Gesture Recognition (FG 2020)}. IEEE, 2020, pp. 734--741.

\bibitem{qu2016cas}
Fangbing Qu, Su-Jing Wang, Wen-Jing Yan, and Xiaolan Fu,
\newblock ``C{AS} ({ME}) 2: A database of spontaneous macro-expressions and
  micro-expressions,''
\newblock in {\em International Conference on Human-Computer Interaction}.
  Springer, 2016, pp. 48--59.

\bibitem{davison2016samm}
Adrian~K Davison, Cliff Lansley, Nicholas Costen, Kevin Tan, and Moi~Hoon Yap,
\newblock ``Samm: A spontaneous micro-facial movement dataset,''
\newblock {\em IEEE transactions on affective computing}, vol. 9, no. 1, pp.
  116--129, 2016.

\bibitem{davison2018objective}
Adrian~K Davison, Walied Merghani, and Moi~Hoon Yap,
\newblock ``Objective classes for micro-facial expression recognition,''
\newblock {\em Journal of imaging}, vol. 4, no. 10, pp. 119, 2018.

\bibitem{yap2020samm}
Chuin~Hong Yap, Connah Kendrick, and Moi~Hoon Yap,
\newblock ``{SAMM} long videos: A spontaneous facial micro-and
  macro-expressions dataset,''
\newblock in {\em 2020 15th IEEE International Conference on Automatic Face and
  Gesture Recognition (FG 2020)}. IEEE, 2020, pp. 771--776.

\bibitem{li2019spotting}
Jingting Li, Catherine Soladie, Renaud S{\'e}guier, Su-Jing Wang, and Moi~Hoon
  Yap,
\newblock ``Spotting micro-expressions on long videos sequences,''
\newblock in {\em 2019 14th IEEE International Conference on Automatic Face \&
  Gesture Recognition (FG 2019)}. IEEE, 2019, pp. 1--5.

\end{thebibliography}

\end{document}